\documentclass{article}

\usepackage[preprint,nonatbib]{neurips_2024}

\usepackage[numbers,square]{natbib}

\usepackage[utf8]{inputenc} 
\usepackage[T1]{fontenc}    
\usepackage{hyperref}       
\usepackage{url}            
\usepackage{booktabs}       
\usepackage{amsfonts}       
\usepackage{nicefrac}       
\usepackage{microtype}      
\usepackage{xcolor}         
\usepackage{graphicx}       
\usepackage{amsmath}

\title{Exploring Large Language Models for Word Games: Who is the Spy?}

\author{Chentian Wei \\
  Institute for Network Sciences and Cyberspace\\
   Tsinghua University\\
  Beijing, China 100084 \\
  \texttt{wct24@mails.tsinghua.edu.cn}\And
   Jiewei Chen \\
  School of Software\\
   Tsinghua University\\
  Beijing, China 100084 \\
   \texttt{chen-jw24@mails.tsinghua.edu.cn} \\  
   \AND
   Jinzhu Xu \\
  Department of Computer Science and Technology\\
   Tsinghua University\\
  Beijing, China 100084 \\
   \texttt{xujz23@mails.tsinghua.edu.cn} \\
}

\begin{document}

\maketitle

\begin{abstract}
Word games hold significant research value for natural language processing (NLP), game theory, and related fields due to their rule-based and situational nature. This study explores how large language models (LLMs) can be effectively involved in word games and proposes a training-free framework. "Shei Shi Wo Di" or "Who is the Spy" in English, is a classic word game. Using this game as an example, we introduce a Chain-of-Thought (CoT)-based scheduling framework to enable LLMs to achieve excellent performance in tasks such as inferring role words and disguising their identities. We evaluate the framework's performance based on game success rates and the accuracy of the LLM agents' analytical results. Experimental results affirm the framework's effectiveness, demonstrating notable improvements in LLM performance across multiple datasets. This work highlights the potential of LLMs in mastering situational reasoning and social interactions within structured game environments.
Our code is publicly available at
\href{https://github.com/ct-wei/Who-is-The-Spy}{https://github.com/ct-wei/Who-is-The-Spy}.

\end{abstract}

\section{Introduction}

\label{gen_inst}

LLMs have made significant progress in solving reasoning and planning problems. Researchers have widely applied these models to tasks like board games and social deduction games, achieving remarkable success. In board games, Noever et al.\cite{The-Chess-Transformer} fine-tuned the GPT-2 model using Portable Game Notation (PGN), optimizing 774 million parameters. This enabled the model to generate reasonable strategies and exhibit game patterns recognizable as classic openings. ChessGPT\cite{feng2024chessgpt} combines strategy learning and language modeling, utilizing a large dataset of chess games to enhance the model's ability to solve complex chess positions. In social deduction games, Xu et al.\cite{xu2023exploring} proposed a framework that allows LLMs to participate in these games without fine-tuning. By reviewing historical dialogues, the model improves its reasoning abilities, making it applicable to social deduction games that rely on natural language interaction. Xu et al.\cite{pmlr-v235-xu24ad} also developed strategic language agents using reinforcement learning (RL) combined with LLMs for games like "Werewolf." These agents can generate diverse actions during the game and choose the best option from multiple candidates, achieving near-human-level strategic play. Wu et al.\cite{wu2024enhance} introduced the "Thinker" module, which processes player speech with structured analysis and deep logical evaluation, enhancing the reasoning and decision-making capabilities of LLMs in games like "Werewolf," where complex language reasoning is required.

Studies (e.g.,  \cite{hao-etal-2023-reasoning,  xiong2024deliberatereasoningllmsstructureaware}) have shown that combining LLMs with world models enhances their planning and reasoning capabilities by generating action plans and predicting future states, leading to strong performance in tasks like logical reasoning, math, and coding. Ku et al.\cite{wang-etal-2024-rethinking-bounds} explored how multi-agent discussion frameworks enhance LLM reasoning, finding that even single-agent setups can achieve comparable results in collaborative reasoning tasks through strong prompting. CR-Planner\cite{li2024elicitreasoningllmscriticguided} combines retrieval-augmented generation and critic models to improve reasoning and planning abilities.

In recent years, Chain of Thought (CoT)\cite{wang2022self} and Tree of Thought (ToT)\cite{yao2024tree} have achieved significant success in enhancing LLM reasoning and planning capabilities. Kim et al.\cite{kim2023cot} demonstrated how CoT fine-tuning improves LLM performance in few-shot learning tasks, significantly boosting zero-shot and few-shot learning abilities in reasoning tasks by fine-tuning CoT datasets. Zhang et al.\cite{zhang2024chain} combined CoT and ToT with preference optimization to enhance reasoning. The method generates multiple candidate paths during reasoning and evaluates each step, optimizing the entire reasoning chain and further improving reasoning quality. This research aims to leverage CoT and ToT to improve LLM performance in reasoning tasks, ultimately enhancing LLM reasoning capabilities in social deduction games like "Who is the Spy."

In this work, we attempted to use agents based on large language models to participate in a classic word game "Who is the spy". In this game, as an agent who is the spy, it needs to determine whether it is the spy or not and try to disguise itself. Meanwhile, as a member who is not the spy, one should avoid giving hints to the spy and then try to guess one's own keyword. At the same time, members who are not the spy also need to accurately infer which player is the spy member. 
Our main contributions are as follows:

\begin{itemize}

\item We have constructed a framework for large language models to play word games, enabling the models to significantly improve the performance of word games without any training and human intervention.

\item We conducted large-scale tests in the context of the classic word game "Who's the spy", and carried out ablation experiments on different modules to verify the effectiveness of the overall framework.

\item In response to special circumstances in the experiments, we conducted targeted case analyses to achieve the interpretability of the language models using the framework. 

\end{itemize}

\section{Related Works}
The emergence of communication among agents in word games has raised significant attention in the research community. This section discusses prior studies that align with our focus, particularly on LLMs in reasoning tasks, Chain-of-Thought (CoT) prompting, social deduction games, and structured multi-agent interactions

\subsection{Large Language Models in Reasoning and Interactive Tasks}
Large Language Models (LLMs) have shown great potential in improving reasoning capabilities, particularly through techniques such as Chain-of-Thought (CoT) prompting. Wei et al. (2022) \cite{wei2022chain} introduced CoT prompting to enhance LLMs' reasoning abilities by encouraging step-by-step logical inference. This approach has been instrumental in solving complex tasks, such as arithmetic reasoning and logical problem-solving. The structured reasoning inherent in CoT is also critical for applications like word games, where roles must be inferred through logic and deduction, as seen in games like "Who is the Spy." Similarly, Kojima et al. (2022) \cite{kojima2022large} extended this idea by showing that LLMs can perform zero-shot reasoning with CoT prompting, highlighting their ability to reason effectively without additional task-specific training.

The emergent reasoning capabilities of modern LLMs have also been explored in the context of more dynamic tasks. Bubeck et al. (2023) \cite{bubeck2023sparks} demonstrated the ability of models like GPT-4 to handle multi-turn reasoning and social inference, both of which are crucial in deduction games. Brown et al. (2020) \cite{brown2020language}, in their foundational work on GPT-3, emphasized how large-scale language models can be applied to interactive tasks, including role-playing and dynamic decision-making scenarios. These capabilities make LLMs particularly well-suited for social deduction games, where players must reason through multiple interactions to identify hidden roles.

Plaat et al. (2024) \cite{plaat2024reasoning} introduced a comprehensive survey on reasoning with LLMs, detailing how these models handle logical inference and multi-step problem-solving. They underscored the importance of reasoning across multiple interactions, which is directly applicable to games that require deduction and strategic thinking. Additionally, Huang and Chang (2023)\cite{huang2023reasoning} focused on the challenges LLMs face, such as maintaining logical consistency and accuracy over extended interactions—issues that are particularly relevant in games requiring role-based reasoning and strategic deduction. In line with this, Ho et al. (2023) \cite{ho2023reasoning} explored how LLMs can serve as reasoning guides, demonstrating that these models can assist both human players and AI agents in navigating complex logical tasks, an essential function in social deduction games like "Who is the Spy."

\subsection{Word Games and LLM Applications}
The application of LLMs in word games, an emerging area of research, offers unique opportunities to explore their reasoning and interactive capabilities. Word games provide a structured environment where LLMs can demonstrate both logical deduction and role-playing abilities. Ouyang et al. (2022) \cite{ouyang2022training} highlighted LLMs' capacity to generate coherent responses in complex interaction scenarios, including those that require deductive reasoning. This is directly applicable to games like "Who is the Spy," where LLMs are tasked with inferring hidden identities based on logical deduction.

Xiao and Yang (2024) \cite{XiaoYang2024} examined the use of LLMs in measuring game difficulty, showing that LLMs can act as game testers to assess the difficulty of game tasks. Their findings suggest that LLMs, although not yet human-level players, provide valuable feedback on game design and balance. This insight is particularly relevant for role-based deduction games, where LLMs can simulate various game scenarios to evaluate the difficulty of detecting the spy through player interactions. 

Merino et al. (2024) \cite{MerinoEarleSudhakaran2024} demonstrated the potential of LLMs in generating word puzzles for The New York Times' Connections Word Game, showcasing how these models can leverage pattern recognition and logical deduction to create engaging and complex puzzles. The same principles can be applied to generating dynamic roles or game scenarios in social deduction games, automating the creation of challenging game content.

Lan et al. (2023) \cite{lan2023llm} explored the use of LLMs in the social deduction game Avalon, highlighting how these models can navigate collaboration and confrontation within the game's dynamic environment. Their work showed that LLMs can simulate social interactions and adapt their reasoning strategies based on the game's evolving context. This ability to reason in varied social settings positions LLMs as powerful tools for role-playing and deduction in multiplayer games. Further, Lan et al. (2023) \cite{xu2023exploring} demonstrated that LLMs can effectively engage in both cooperative and adversarial strategies in Avalon, underscoring the versatility of LLMs in complex, dynamic environments.

\subsection{Chain of Thought (CoT) Prompting in Small-Scale Games}
CoT\cite{wang2022self} prompting plays a crucial role in small-scale games, where it helps intelligent agents reason and act in a controlled, language-based game environment. By observing the prompts or statements provided by players step by step, agents engage in logical reasoning and decision-making. CoT enables agents to break down complex scenarios, evaluate the game's context, predict potential outcomes, and adjust their strategies accordingly. This step-by-step reasoning helps agents avoid biases and make human-like decisions, crucial for games like "Who is the Spy."

In "Who is the Spy", CoT prompting divides agents into two camps: villagers and spies. The agents, using CoT, simulate logical replies, maintain consistency, and predict the reactions of other players. This enables spies to mislead others by offering clues that align with the group's expectations, while also making it harder for others to detect their identity. On the other hand, villagers use CoT to identify inconsistencies in other players' statements. By breaking down each statement systematically, villagers can find contradictions and increase their chances of identifying the spy. In addition, CoT allows players to ask strategically framed questions, making it easier to uncover flaws in the spy’s reasoning and expose their true identity.

Moreover, Kojima et al. (2022) \cite{kojima2022large} demonstrated that CoT prompting enables LLMs to reason without explicit task-specific training, an essential insight for implementing flexible, training-free frameworks. This ability to reason through step-by-step logic is critical for multi-agent environments, as it allows LLMs to collaborate, deceive, and deduce within dynamic game scenarios, as discussed throughout this paper.

\subsection{Recent Advances in Multi-Agent Reasoning for Word Games}
While LLMs have shown considerable potential in reasoning tasks, challenges remain in applying them to long-horizon tasks and maintaining accuracy over extended interactions. As highlighted by Li et al. (2023) \cite{li2023theory}, overcoming these challenges requires innovative approaches such as belief state representations to mitigate issues like hallucination and improve collaborative decision-making. Our study aims to build on these insights to further explore the capabilities of LLMs in multi-agent word games, particularly those involving role-based reasoning, social interaction, and deductive reasoning. By examining how LLMs can maintain accuracy and consistency across long interactions, we hope to enhance their application in dynamic, real-time game environments.

\section{Definition}
\label{headings}

\begin{figure}[t]
\centering
\includegraphics[width=0.8 \linewidth]{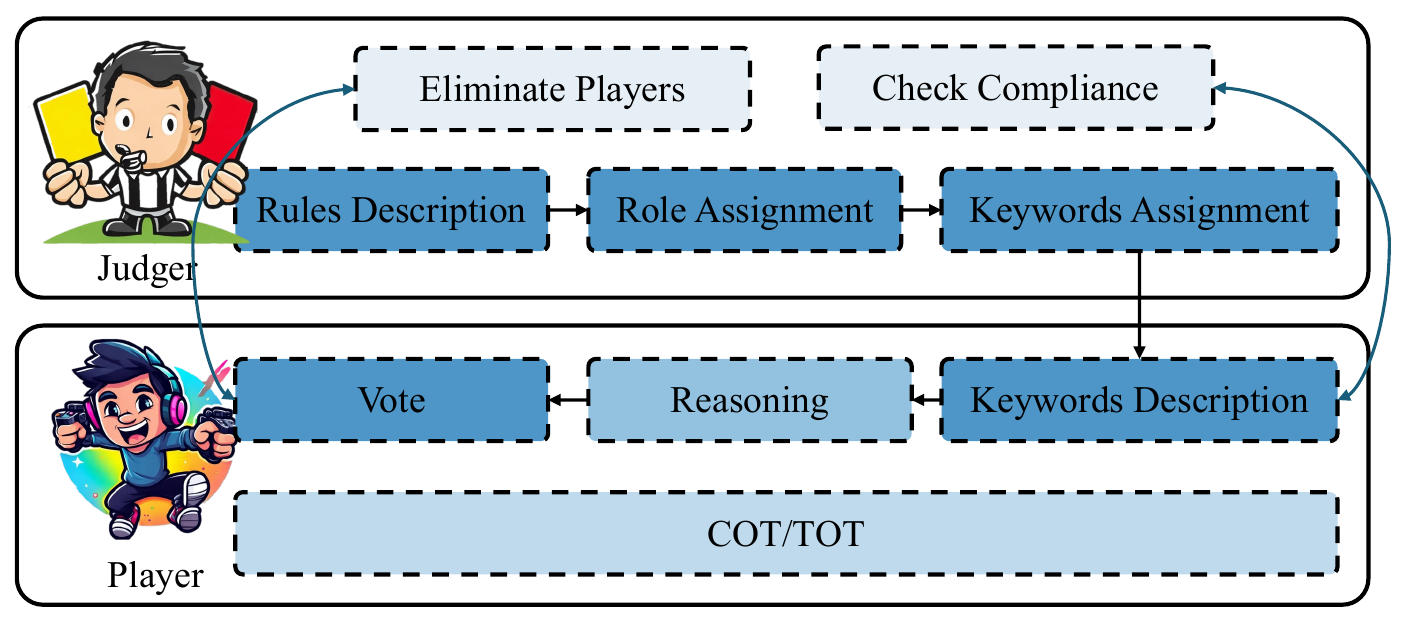}
\caption{Our Framework} 
\label{framework}
\end{figure}

"Who is the Spy" is a highly popular party game in China where players must use dialogue and reasoning to identify the spy or antagonist hidden within the group. The core gameplay of this game shares similarities with "Werewolf" and "Mafia," as all three revolve around hidden identities, deception, and deduction.

The process of the game is as follows: 
\begin{enumerate} 
    \item Rule Description: Explain the game rules to the large language model, including the description of keywords, player elimination, and voting procedures. 
    \item Role Assignment: Assign each player an initial role, which could be either the spy or a villager. Ensure that the model is aware of its assigned role. 
    \item Keyword Assignment: Assign a keyword to each role in the game. One is the spy's keyword, while the others are similar but not identical to the spy's keyword. Inform the large language model of the keyword it has been assigned. 
    \item Keyword Description: Players take turns describing the keyword they have received, and the large language model describes its assigned keyword based on the given information. \item Vote: At the end of this round, all players eliminate the non-spy players and vote to select the player who is most likely to be the spy. 
\end{enumerate}

\begin{figure}[t]
\centering
\includegraphics[width=\linewidth]{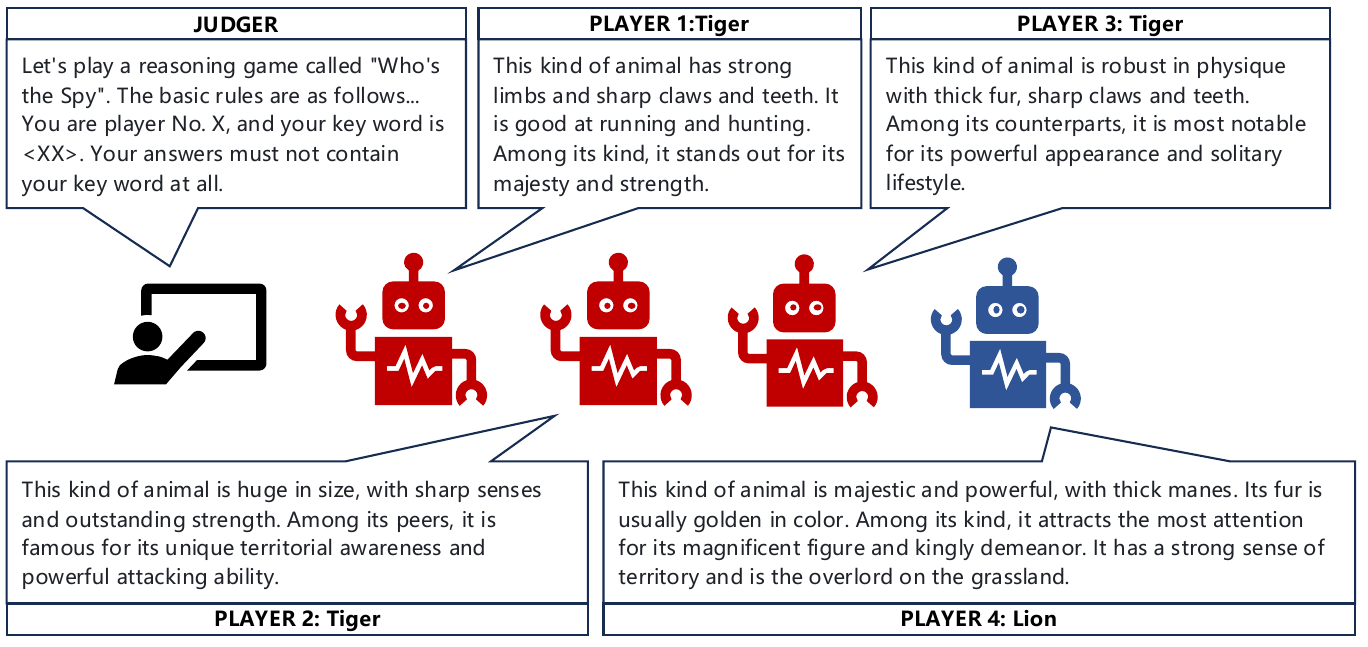}
\caption{The game process of Who is the Spy} 
\label{framework}
\end{figure}

Based on the structure described above and as shown in Figure \ref{framework}, we divide the overall framework into two parts: the referee and the players.

In the game, the referee controls the progression of the game. First, the referee explains the game rules, assigns roles to all players, and distributes key terms. During the game session, the referee continuously checks whether the players' outputs comply with the rules. If an output does not adhere to the game rules, the player will be required to submit a new output, and the type of violation will be recorded. Next, the referee compiles all role descriptions and distributes them uniformly to the players for reasoning and voting. The referee is responsible for collecting the voting results and carrying out the elimination process. Both large language models and humans may participate in the game simultaneously, and they will play according to the rules.

\section{Method}

\subsection{Overview}

\begin{figure}[b]
\centering
\includegraphics[width=\linewidth]{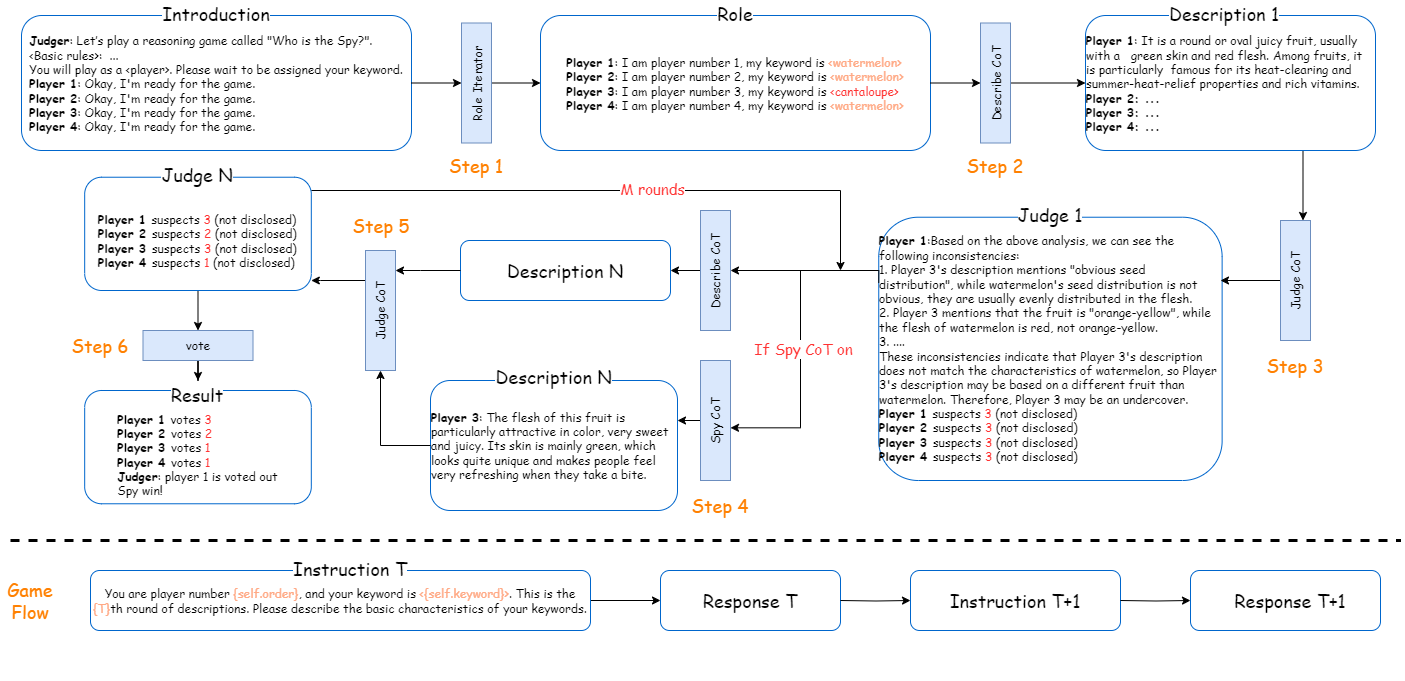}
\caption{Overview of Game Flow} 
\label{overview}
\end{figure}

The overall game process is shown in Figure \ref{overview}. First, the judge introduces the rules of the game to the four LLM agents and ensures they understand. Then, the Role Iterator assigns a player role to each LLM. After each player receives their keyword, they will provide a description, which is generated by Describe CoT. Next, all players will use Judge CoT to make spy judgments based on the first round of descriptions. These judgments are not made public. In the subsequent M rounds, the four players will repeat the description and judgment process. The difference is that when the Spy CoT option is activated, players who identify themselves as spies will use Spy CoT to conceal their spy identities. Finally, after completing multiple rounds of judgment, all players will vote based on their final judgment results, and the system will determine the winner of the game.

\subsection{Role Iterator}
The role assigner specifies the role that each LLM plays in the game. 

Definition: 

\begin{itemize}
    \item The set of players is $P = \{p_1, p_2, p_3, p_4\}$.
    \item The set of civilian words is $W_c = \{w, w, w\}$, and the spy word is $W_u = \{w_u\}$.
    \item $w$ and $w_u$ are two semantically similar but different words, satisfying $\text{Sim}(w, w_u) > \theta$ (similarity threshold).
    
\end{itemize}

The allocation formula is as follows: 

Define the player's word as $R: P \to \{w, w_u\}$, where: 

\begin{equation*} 
    R(p_i) = 
    \begin{cases} 
        w & \text{if }\ p_i\ \text{civilian}  \\ 
        w_u & \text{if }\ p_i\ \text{spy} 
    \end{cases}
\end{equation*}

The following formula is satisfied:
\begin{equation*}
    \begin{aligned}
        |\{p_i \mid R(p_i) = w\}| &= 3 \\
        |\{p_i \mid R(p_i) = w_u\}| &= 1
    \end{aligned}
\end{equation*}

\subsection{Describe Phase}

In each description phase, each player $p_i$ gives a description $D_i$ based on his word $R(p_i)$, and the description set is $D = \{D_1, D_2, D_3, D_4\}$.The description process can be expressed as the following formula:
\begin{equation*}
    \begin{aligned}
        D_i =(R(p_i))
    \end{aligned}
\end{equation*}
Where\ $ \text{CoT\_Describe}: \{w, w_u\} \to \text{Description Space}$. 
 
\begin{figure}[t]
\centering
\includegraphics[width=\linewidth]{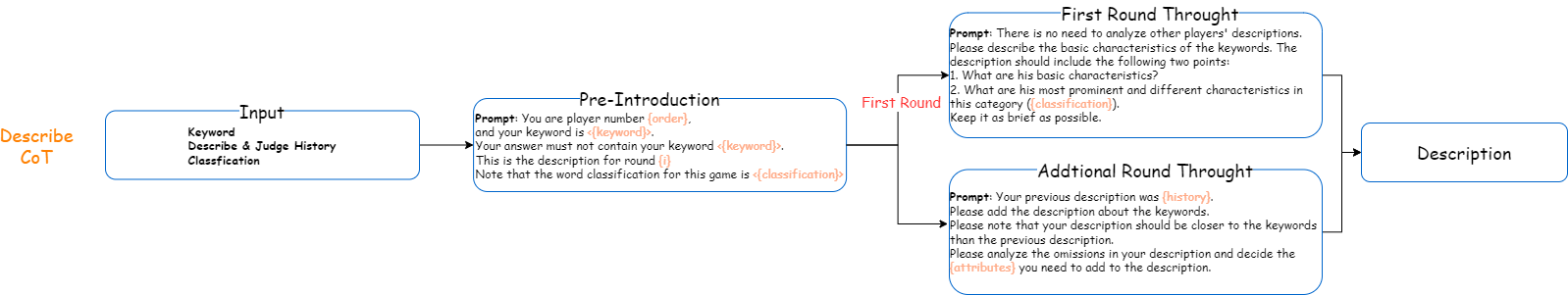}
\caption{CoT of Description} 
\label{describe}
\end{figure}

 We adopt the design shown in the figure \ref{describe} for the described CoT. CoT inputs the keywords held by the player, the history of description and judgment, and the word classification used for model judgment. In the pre-introduction stage, the model is first made to understand its player information, game situation and important output rules.
 
In the first round, CoT guides the model to output the most basic information and the most prominent features of the keywords in the word classification, facilitating more accurate judgments.

In the description of subsequent rounds, the CoT prompt model uses the previous description to supplement. In the subsequent description, the model will focus on describing the special attributes of the keyword that are not mentioned by other players. At the same time, to avoid the model exposing too much information about the keyword, CoT limits the number of words output by the model. Our experiments have proved that such CoT design is effective.

\subsection{Judge Phase}

In the judgment phase, each player $p_i$ judges the words $R(p_j)$ of all players $p_j$ based on the descriptions $D_j$ of other players and his own words $R(p_i)$.

The judgment process can be expressed as a formula of the form $J_i: D \to \{w, w_u\}^4$:

\begin{equation*}
    \begin{aligned}
        J_i(D) = \{ \hat{R}(p_j)=\text{CoT\_Judge}(p_j, D_j) \mid \hat{R}(p_j) \in \{w, w_u\}, p_j \in P\}
    \end{aligned}
\end{equation*}

Where $\hat{R}(p_j) $ is $p_i $’s guess about the word $p_j $. 

\begin{figure}[b]
\centering
\includegraphics[width=\linewidth]{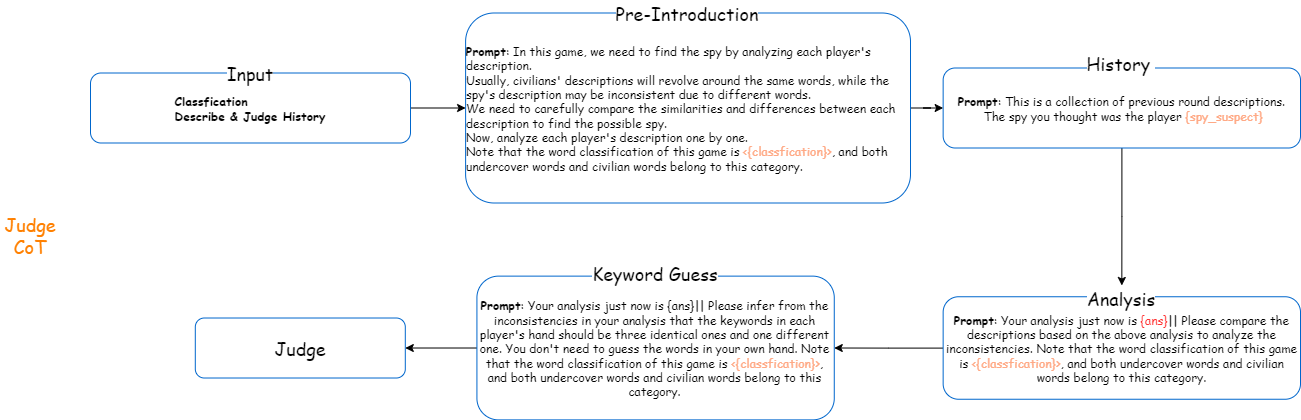}
\caption{CoT of Judgement} 
\label{judge}
\end{figure}

The overall process of judging CoT is shown in Figure \ref{judge}. The input consists of the entire description history of the four players and word classification. We do not input the keywords held by the players themselves, as this helps the model make more unbiased word judgments.

Similar to the process described above, CoT also begins with a pre-introduction. It introduces the basic reasoning logic of humans in spy judgment to the model and emphasizes that spy words also belong to the same category. For rounds after the first, we introduce the previous inference history and the spy judgments that have been made, which aligns with the human thinking process.

In the judge phase, we believe that having the model guess the keywords held by the four players directly, according to the rules of the game, is better than directly inferring the spy. On one hand, this approach mirrors human thinking. On the other hand, it helps mitigate the illusion phenomenon that the model might produce, such as drawing conclusions that contradict subsequent analyses or mistakenly identifying multiple spies. This method also encourages the model to consider the game situation more comprehensively.

Finally, the model will output the keywords it has judged. Experiments show that the error rate of simple models is relatively high when they simply output the words they have judged without reasoning. To address this, we use an external agent method, allowing the model to output a specific format. The program then evaluates the spy the model has chosen.

\subsection{Vote}

During the voting phase, the model will vote according to the final judgment. We have the following definition:

The voting matrix:
\begin{equation*}
    \begin{aligned}
        V \in \mathbb{N}^{4 \times 4}
    \end{aligned}
\end{equation*}

Where $V_{ij} $represents the number of votes cast by player $p_i$ for player $p_j $.

Voting rules:

\begin{itemize}
    \item Players cannot vote for themselves: $V_{ii} = 0, \forall i$.
    \item If player $p_i$ suspects that he is a spy agent, randomly select $p_k (k \neq i)$ to vote.
    
\end{itemize}

The voting decision of player $p_i $is: 
\begin{equation*}
    \begin{aligned}
        v_i = \arg\max_{p_k \in P \setminus \{p_i\}} \text{Conf}(p_k, J_i(D))
    \end{aligned}
\end{equation*}

Where $\text{Conf}(p_k, J_i(D))$ represents the confidence of $p_i$ that $p_k$ is a spy. Currently, we use a simplified version, that is, the confidence of a single player is 1, and the confidence of the other two positions is 0.

The total number of votes of player $p_k$ is:

\begin{equation*}
    \begin{aligned}
         V_k = \sum_{i=1}^4 V_{ik}.
    \end{aligned}
\end{equation*}

The final victory or defeat of the system is determined by the player with the most votes being $p_t$, which satisfies:
\begin{equation*}
    \begin{aligned}
         p_t = \arg\max_{p_k \in P} V_k
    \end{aligned}
\end{equation*}

Winning and losing rules:

\begin{itemize}
    \item If $p_t$ is a spy player, the civilian wins.
    \item If $p_t$ is not a spy player, the spy wins.
    
\end{itemize}

Formula:

\begin{equation*}
    \begin{aligned}
        \text{Result} = \begin{cases} \text{Civilian wins} & \text{If } R(p_t) = w_u, \\ \text{Spy wins} & \text{If } R(p_t) \neq w_u. \end{cases}
    \end{aligned}
\end{equation*}

\begin{figure}[b]
\centering
\includegraphics[width=\linewidth]{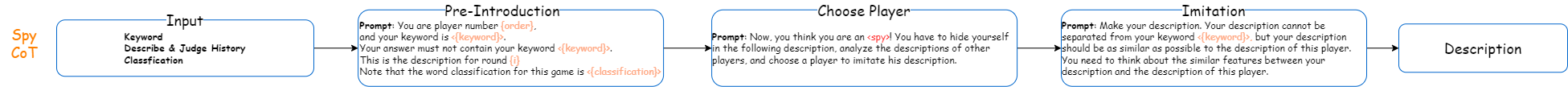}
\caption{CoT of Spy} 
\label{spy}
\end{figure}

\subsection{Spy Disguise}

In addition, after CoT greatly improved the civilians' winning rate, we introduced CoT for spies to disguise themselves. The process is shown in Figure \ref{spy}.

After the first round of judgment, the player who identifies as a spy will use the spy CoT. The input for this CoT is the same as the description CoT, and the pre-introduction process is also identical to that of the description CoT.

In the rules of this game, the spy only needs to make at least one player choose the same role as themselves to win the game. Based on this observation, CoT will instruct the model to find a civilian player who is easy to disguise, and whose description does not obviously differ from the spy's own keywords. The model will then judge the civilian's keywords and output features similar to its own, thereby disrupting the judgment of the civilians.

In subsequent experiments, this CoT has shown some effectiveness. However, due to multiple rounds of judgment, spies often end up identifying themselves as non-spies in later rounds. Civilians may also mistakenly identify themselves as spies. At the same time, we lack effective experiments to prove that the spy's output does not deviate from its own keywords, which requires further investigation in future research.

\section{Experiment}

\subsection{Implementation Details}

We developed the “Who is Spy?” game program in Python. We used Zhipu AI's GLM4-9b-Flash as our baseline model. In all experiments, we set the temperature of the proxy model to 0.3. The maximum number of tokens is 10,000. The original dataset used by Role Iterator comes from OpenAI's o1 model, and o1 performs word similarity judgment. Our dataset has one hundred different word groups.

\subsection{Evaluation}

We conducted 100 experiments using our framework for three cases: without CoT, with description CoT and judgment CoT, and with description CoT, judgment CoT, and spy CoT. For all models, our system prompt is: "You are not an AI assistant now, you are a player participating in the game 'Who is the Spy.' Please output content that matches the player's identity."

For the case without CoT, our description prompt is: "You are player No. <order>, and the keyword you are assigned is: <keyword>. Please start describing. Only output the description, and do not include any other information. Note that your answer must not contain your keyword <keyword>." The judgment prompt is: "Please vote and enter the number of the player you think is the spy. This number should be between 1 and 4. Do not output any other information."

For the case without CoT, we also explained the rules of the game to the model, along with all previous description history.

\subsubsection{Evaluation Metrics}

Finally, we evaluate the performance of our framework based on metrics from two perspectives.

From the perspectives of game results and strategy, we use metrics related to the gameplay outcome and strategies to quantitatively assess the performance of the proposed agents and the baseline agents.

\textbf{Civilian Winning Rate (CWR).} The civilian winning rate is the percentage of games won by civilians out of the total games played. It is calculated by dividing the number of wins by the total number of games played:

\begin{equation*}
    \begin{aligned}
        CWR = \left( \frac{\#Civilian\ Wins}{\#Games\ Played} \right) \times 100\%
    \end{aligned}
\end{equation*}

\textbf{Civilian Miss Rate (CMR).} The civilian miss rate is the percentage of incorrect votes cast by civilians out of all civilian votes. It is calculated by dividing the number of incorrect civilian votes by the total number of civilian votes:

\begin{equation*}
    \begin{aligned}
        CMR = \left( \frac{\#Civilian\ Wrong\ voting}{\#Civilian\ voting} \right) \times 100\%
    \end{aligned}
\end{equation*}

From the behavioral perspective, we use a case study approach to analyze whether the LLM performs better in specific cases under our framework. Our analysis is primarily based on manual evaluation, assisted by OpenAI's GPT-4. We focus on the following features:

\begin{itemize}
    \item With the help of CoT, is the model's description of keywords more accurate?
    \item With the help of CoT, is the model's judgment and reasoning more logical?
    \item With the help of CoT, did the spy agent learn how to hide himself without breaking the rules?
    
\end{itemize}

\subsubsection{Game Output}

We conducted experiments on the baseline and our three different benchmarks. The experimental results are shown in the table \ref{exper}.

\begin{table}[!ht]
    \caption{Experiment Result (NC is No CoT. JC is Judge CoT. DC is Describe CoT. SC is Spy CoT)}
  \label{exper}
    \centering
    \begin{tabular}{lllll}
    \hline
        \toprule
        \textbf{Method} &\textbf{ NC(Baseline)} & \textbf{JC} & \textbf{JC \& DC} & \textbf{JC \& DC \& SC}  \\ 
        \midrule
       \textbf{ Game Count} & 100 & 100 & 100 & 100 \\ 
        \textbf{Spy Out} & 7 & 72 & 81 & 75  \\ 
      \textbf{Civilian Out} & 68 & 13 & 15 & 22 \\ 
        \textbf{Draw} & 25 & 15 & 4 & 3  \\ 
        \textbf{CWR} & 7\% & 72\% & 81\% & 75\%  \\ 
        \textbf{CMR} & 66.7\% & 20\% & 16.7\% & 25\%  \\ 
        \bottomrule
    \end{tabular}
\end{table}

According to our experimental results, it is proved that our method has a certain effect on improving the model's evaluation indicators of game results.

For the experimental baseline, our results show that the most basic prompt makes it difficult for civilians to win. Importantly, this result does not imply that the spy agent performs well in the baseline; rather, the large number of civilians makes it easier for the model to vote without reasoning. In fact, in our subsequent case analysis, we observed that the votes output by the model in the baseline case were entirely meaningless and essentially random numbers.

From the experimental results, it is clear that Judge CoT plays a crucial role. It enables the model to vote for spy agents with high accuracy even under simple descriptions. Judge CoT introduces reasoning capabilities to the model, allowing it to logically deduce answers rather than vote randomly. However, when using Judge CoT alone, due to the uncertainty in the description output, we found that incorrect descriptions can lead the model to make incorrect conclusions. This error is not related to the inference process itself but rather to the quality of the descriptions.

The experimental results also show that Describe CoT has a positive effect on improving the civilian win rate. It addresses the issue of hallucinations that occur in the case of a single prompt, leading to incorrect descriptions. In our subsequent case studies, we found that Describe CoT helps the model identify issues in its previous descriptions and self-correct them. Additionally, the model’s descriptions became step-by-step, avoiding the irrelevant output seen in the baseline, which is exactly what we aimed for.

\begin{figure}[b]
\centering
\includegraphics[width=\linewidth]{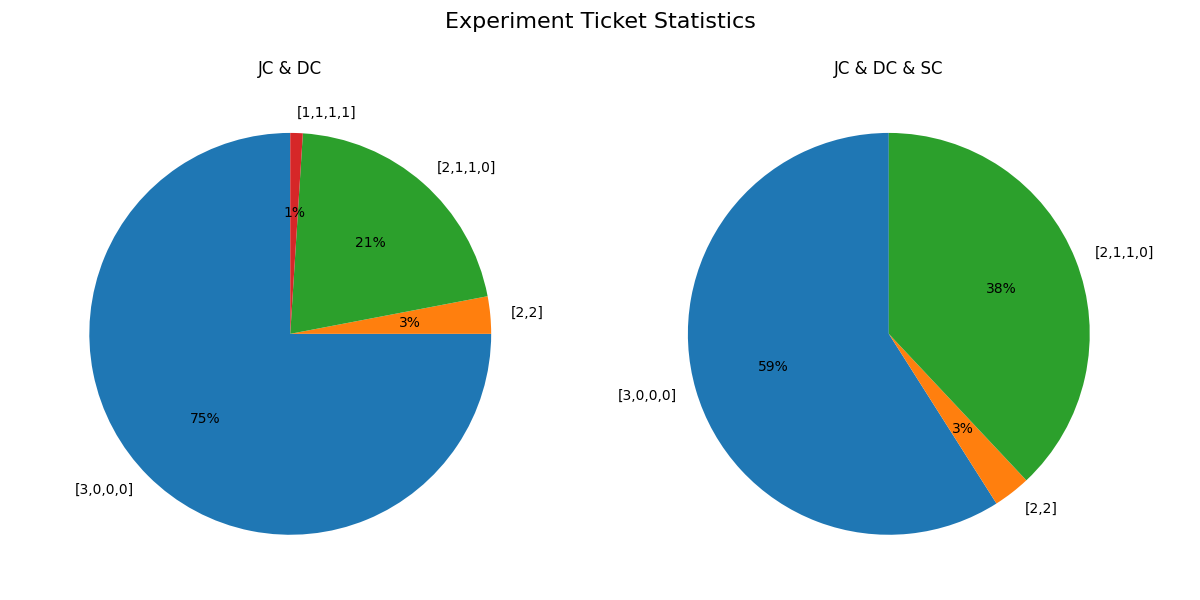}
\caption{Experiment Ticket Statistics} 
\label{ticket}
\end{figure}

In fact, in our numerous experiments, Spy CoT did not significantly improve the spy win rate. However, we found that introducing Spy CoT can increase the probability of civilians voting incorrectly. As shown in the figure \ref{ticket}, our vote statistics indicate that Spy CoT can guide at least one civilian to cast a wrong vote, thereby increasing the likelihood of a tie. However, the introduction of Spy CoT does cause some confusion in the game. For instance, after the second round of judgment, the spy player might mistakenly believe that they are not a spy and induce civilians to think that they are. The specific design of Spy CoT is an area that requires further exploration and discussion.

\subsubsection{Case Study}

First, let’s compare the No CoT and Judge CoT cases. In the example of ['bear', 'bear', 'bear', \textbf{'lion'}], player 3 gave different responses to the same description input. The descriptions of the four players are as follows: "Player 1 said: A common animal in the forest, furry, likes to climb trees. Player 2 said: There is an animal living in the forest, it has thick fur, always likes to climb trees, looks very cute. Player 3 said: The big and cute guy in the forest, it likes to shuttle between trees, and sometimes goes to the river to play. Player 4 said: The majestic animal in the forest, strong and powerful, often strolls on the grassland." In this example, the No CoT case directly gives the answer '2', which is obviously not based on any reasoning, but just outputs a random number from 1 to 4 according to the prompt. The reasoning process guided by Judge CoT is as follows:

"Player 2's description is similar to Player 1's, mentioning forests, thick fur, climbing trees, and cuteness, which also points to bears.Player 3's description uses "cute" and "running through trees", which are consistent with the image of bears, but mentions "playing by the river", which Players 1 and 2 did not mention but may also be bear behavior.Player 4's description mentions "majestic animals" and "walking on the grassland", which is inconsistent with the image of bears. Bears are usually associated with forests and trees, not grasslands."

The final keyword judgment given by the model is "Player 1: Bear, Player 2: Bear, Player 3 (self): Bear, Player 4: Lion". In the end, the model successfully cast spy 4. In this example, we can see the role of Judge CoT in introducing reasoning capabilities to the model.

In this example, we can easily find that the model has hallucinations in its description. For example, the descriptions of players 1, 2, and 3 all mention that bears can climb trees, but this is not common. At the same time, player 4's description is self-contradictory. \textbf{Lions cannot appear in grasslands and forests at the same time}. This shows that the model has a huge flaw in describing keywords in the No CoT case. This is greatly improved in the Describe CoT case. This is the description output by player 4 using Describe CoT in the same game scenario:

\begin{itemize}
    \item "This is a large animal with thick hair and long legs. The most special thing about it among its kind is that it has a unique mane and great strength. "
    \item "This animal lives in arid grasslands, and they usually seek shaded places to avoid the heat during the hot season. Their body structure allows them to run quickly and adapt to the vast grasslands. They feed on meat and are strong among predators. Although they show great strength in battle, they show a gentle side in the family, especially with their cubs." 
    \item "This animal is the overlord of the grasslands. They have a strong neck and sharp teeth. The most notable feature of their kind is their facial expression, which gives people a sense of wisdom and mystery. This animal shows amazing speed and agility when hunting. They chase prey on the grasslands and are the best among predators. Although they are extremely ferocious when hunting, their behavior towards their cubs and when protecting their territory is very gentle."
    
\end{itemize}

We can clearly see that Describe CoT uses a description method that first introduces the description and then adds details. This is significantly better than the case without CoT. Because with the help of CoT, the description is relevant and the model will use the history of the description. More detailed supplementary information is conducive to the victory of civilians. At the same time, this helps to avoid the model's hallucination phenomenon. In the inspection of GPT o1, no phenomenon of incorrect description was found when Describe CoT was used.

For Spy CoT we use the following example. For Spy CoT we use the following example. In the game ['Watermelon', \textbf{'cantaloupe'}, 'Watermelon', 'Watermelon'], Player 2 judges himself as an spy agent in the first round of the game, and then his next round description is:

"The flesh of this fruit is particularly attractive in color, very sweet and juicy. Its skin is mainly green, which looks quite unique and makes people feel very refreshing when they take a bite."

This description is extremely confusing, even for humans it is difficult to tell. It also leads to the final outcome of the game where the spy agent wins. This proves the effectiveness of Spy CoT.

But in some cases, Spy CoT will output overly vague statements, which will lead to confusion in the game. In the game ['plane', \textbf{'car'}, 'plane', 'plane'], Player 2's description in the second round is "It is the most special thing about similar transportation tools in that it is usually designed with seats and can accommodate multiple people at the same time." This overly vague statement made him later judge that he was not an spy agent, while Player 3 judged that he was an spy agent. So Spy CoT needs to be optimized.

\section{Conclusion}
In this study, we explored the application of large language models (LLMs) in the word game "Who is the Spy" and proposed a training-free framework. By introducing a Chain-of-Thought (CoT)-based scheduling framework, we enabled LLMs to achieve excellent performance in tasks such as inferring role words and disguising identities. Through extensive experiments and evaluations, we demonstrated the effectiveness of our framework in improving the performance of LLMs in word games.

The experimental results showed that Judge CoT significantly enhanced the model's reasoning ability, allowing it to make more accurate judgments. Describe CoT effectively addressed the issue of hallucinations and improved the quality of the model's descriptions. Although Spy CoT did not significantly increase the spy win rate, it could increase the probability of civilians voting incorrectly and introduce some confusion into the game.

Our work highlights the potential of LLMs in mastering situational reasoning and social interactions within structured game environments. It also provides valuable insights and a practical framework for future research on the application of LLMs in word games and other similar scenarios. However, there are still some limitations and areas for improvement. For example, the design of Spy CoT needs to be further optimized to avoid confusion and ensure the consistency of the spy's behavior. Additionally, more advanced techniques and strategies could be explored to further enhance the performance and intelligence of LLMs in game-playing.

Overall, this study contributes to the growing body of research on the intersection of LLMs and game applications, opening up new avenues for the development and utilization of intelligent language models in the field of entertainment and beyond. 

\bibliographystyle{plain}
\bibliography{neurips_2024}

\end{document}